\documentclass[letterpaper, 10 pt, conference]{ieeeconf}  
\usepackage[T1]{fontenc}

\IEEEoverridecommandlockouts                              

\usepackage{graphics} 
\usepackage{epsfig} 
\usepackage{mathptmx} 
\usepackage{times} 
\usepackage{amsmath} 
\usepackage{amssymb}  
\usepackage{multirow}
\usepackage{makecell}
\usepackage{cite}
\usepackage{kotex}
\usepackage[dvipsnames]{xcolor}
\usepackage{hyperref}
\usepackage{caption}
\usepackage{booktabs}
\usepackage{adjustbox}

\abovecaptionskip
\belowcaptionskip
\usepackage{indentfirst}

\title{\LARGE \bf
X-MAS: Extremely Large-Scale Multi-Modal Sensor Dataset \\for Outdoor Surveillance in Real Environments 
\vspace{-0.8cm}
}

\author{DongKi Noh$^{1, 2}$, Changki Sung$^{1}$, Teayoung Uhm$^{3}$, WooJu Lee$^{1}$, Hyungtae Lim$^{1}$, Jaeseok Choi$^{4}$, Kyuewang Lee$^{5}$,\\Dasol Hong$^{1}$, Daeho Um$^{5}$, Inseop Chung$^{4}$, Hochul Shin$^{6}$, MinJung Kim$^{7}$, Hyoung-Rock Kim$^{2}$,\\ SeungMin Baek$^{2}$, and Hyun Myung$^{1*}$, \textit{Senior Member, IEEE}
\thanks{*This research was supported in part by Institute for Information \& Communication Technology Promotion (IITP) grant funded by the Korea government (MSIT) (No. 2017-0-00306) and in part by the Korean Evaluation Institute of Industrial Technology (KEIT) funded by the Ministry of Trade Industry and Energy (MOTIE) (No. 10080489)}
\thanks{$^*$Corresponding author: Hyun Myung}
\thanks{$^{1}$DongKi Noh, C. Sung, W. Lee, H. Lim, D. Hong and Hyun Myung are with School of Electrical Engineering at Korea Advanced Institute of Science and Technology (KAIST), Daejeon, 34141, Republic of Korea. \{dongki.noh, cs1032, dnwn24, shapelim, ds.hong, hmyung\}@kaist.ac.kr}%
\thanks{$^{2}$DongKi Noh, Hyoung-Rock Kim and SeungMin Baek are with the Advanced Robotics Lab. at LG Electronics, Seoul, 06772, Republic of Korea  \{dongki.noh, hyoungrock.kim, seungmin2.baek\}@lge.com}%
\thanks{$^{3}$Teayoung Uhm is with the Korea Institute of Robotics and Technology Convergence (KIRO), Pohang, 37666, Republic of Korea. uty@kiro.re.kr}%
\thanks{$^{4}$J.S. Choi and I.S. Chung are with Department of Intelligence and Information, Seoul National University (SNU), Seoul, 08826, Republic of Korea \{jaeseok.choi, jis3613\}@snu.ac.kr}%
\thanks{$^{5}$K.W. Lee and D.H. Um are with Department of Electrical and Computer Engineering, Automation and Systems Research Institute (ASRI), SNU, Seoul, 08826, Republic of Korea \{kyuewang, umdaeho1\}@snu.ac.kr}%
\thanks{$^{6}$Hochul Shin is with the Electronics and Telecommunications Research Institute (ETRI),  Daejeon, 34129, Republic of Korea. creatrix@etri.re.kr}%
\thanks{$^{7}$MinJung Kim is with Kim Jaechul Graduate School of Artificial Intelligence, KAIST, Daejeon, 34141, Republic of Korea. emjay73@kaist.ac.kr}%
\vspace{-0.3cm}
}

\begin{document}

\maketitle
\thispagestyle{empty}
\pagestyle{empty}

\begin{abstract}
In robotics and computer vision communities, extensive studies have been widely conducted regarding surveillance tasks, including human detection, tracking, and motion recognition with a camera. Additionally, deep learning algorithms are widely utilized in the aforementioned tasks as in other computer vision tasks.  
Existing public datasets are insufficient to develop learning-based methods that handle various surveillance for outdoor and extreme situations such as harsh weather and low illuminance conditions. Therefore, we introduce a new large-scale outdoor surveillance dataset named \textit{e\textbf{X}tremely large-scale \textbf{M}ulti-mod\textbf{A}l \textbf{S}ensor dataset} (\textit{X-MAS}) containing more than 500,000 image pairs and the first-person view data annotated by well-trained annotators. Moreover, a single pair contains multi-modal data (e.g. an IR image, an RGB image, a thermal image, a depth image, and a LiDAR scan). This is the first large-scale first-person view outdoor multi-modal dataset focusing on surveillance tasks to the best of our knowledge. We present an overview of the proposed dataset with statistics and present methods of exploiting our dataset with deep learning-based algorithms. The latest information on the dataset and our study are available at \url{https://github.com/lge-robot-navi}, and the dataset will be available for download through a server.


\section{INTRODUCTION}

\end{abstract}
The surveillance robot~\cite{ginting2021chord,zaheer2021anomaly,hoshino2015probabilistic,xu2010systems} is being popular topic
in robotics field with the development of robotics~\cite{chen2019suma++,kim2022step,lim2021patchwork}.
Over the past several decades, researchers have studied indoor/outdoor surveillance algorithms applied to multiple fixed cameras~\cite{bozcan2021gridnet,xu2010systems} and to a mobile robot equipped with a camera~\cite{hoshino2015probabilistic}. In the recent decade, researchers have started focusing on deep-learning methods to deal with harsh outdoor environments~\cite{khaire2022semi,zaheer2021anomaly,gruosso2021human,bozcan2020uav,Rudenko2021Learning}. 
Moreover, recent methods leverage multi-modal sensors because the RGB camera-only methods for outdoor surveillance have an obvious limitation caused by image quality degradation when handling various environmental changes~\cite{khaire2022semi,liu2019ntu,Ye9696362Tracker,lei2006real}.

\begin{figure}[t]
\centering
\includegraphics[width=7.0 cm]{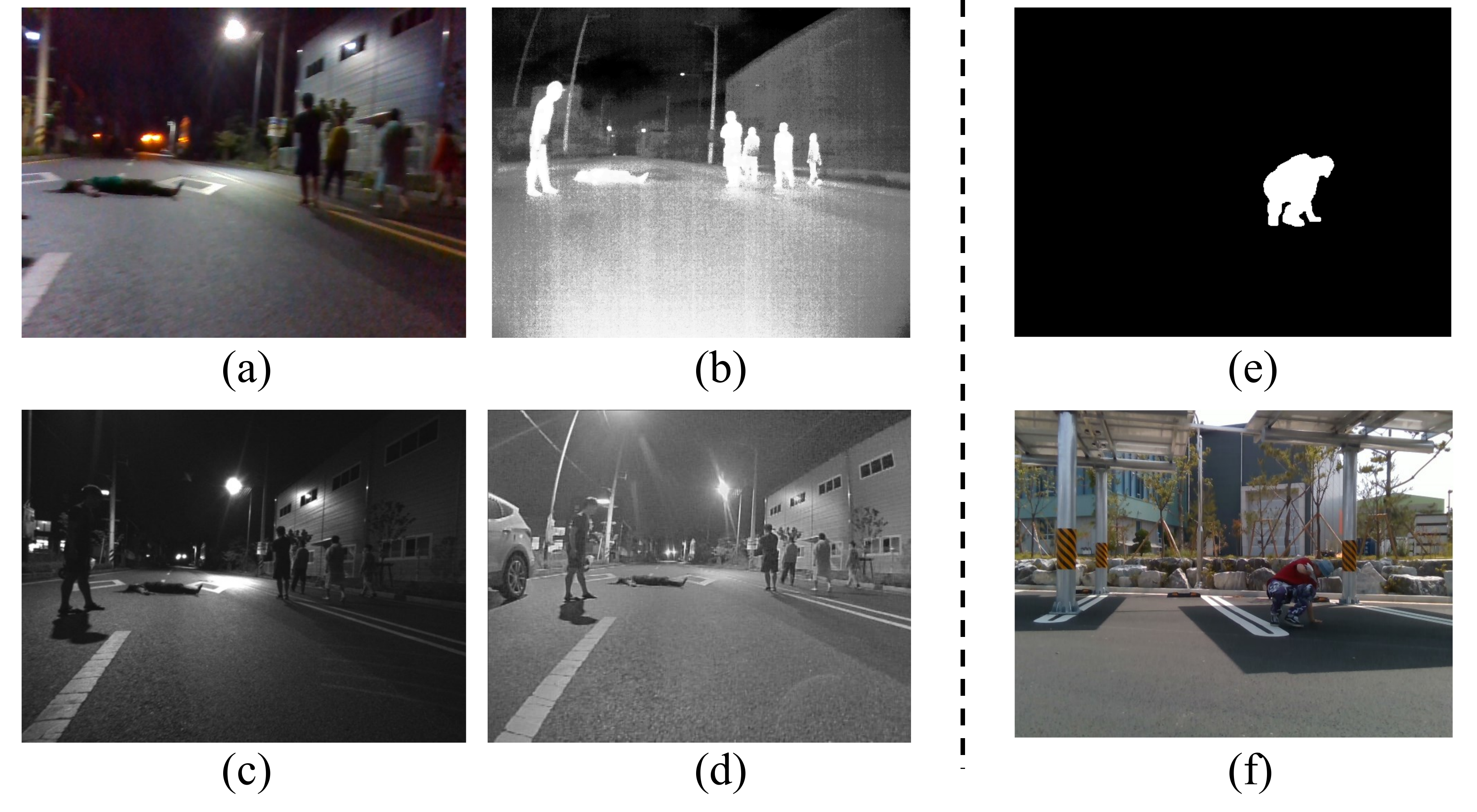}
\captionsetup{font=footnotesize}
\caption{Examples of the first-person view multi-modal dataset: (a) an RGB image, (b) a thermal image, (c) an IR image, and (d) a night vision image. Our dataset provides mask images (e) corresponding to an RGB image (f). It also provides annotations of bounding boxes and tracking IDs. The dataset has been collected for a long period ('$17\sim$'$21$) and consists of more than 2.5 million images. \label{fig_example_of_dataset}}
\vspace{0.25cm}
\end{figure} 

\begin{figure}[t]
\centering
\includegraphics[width=7.0 cm]{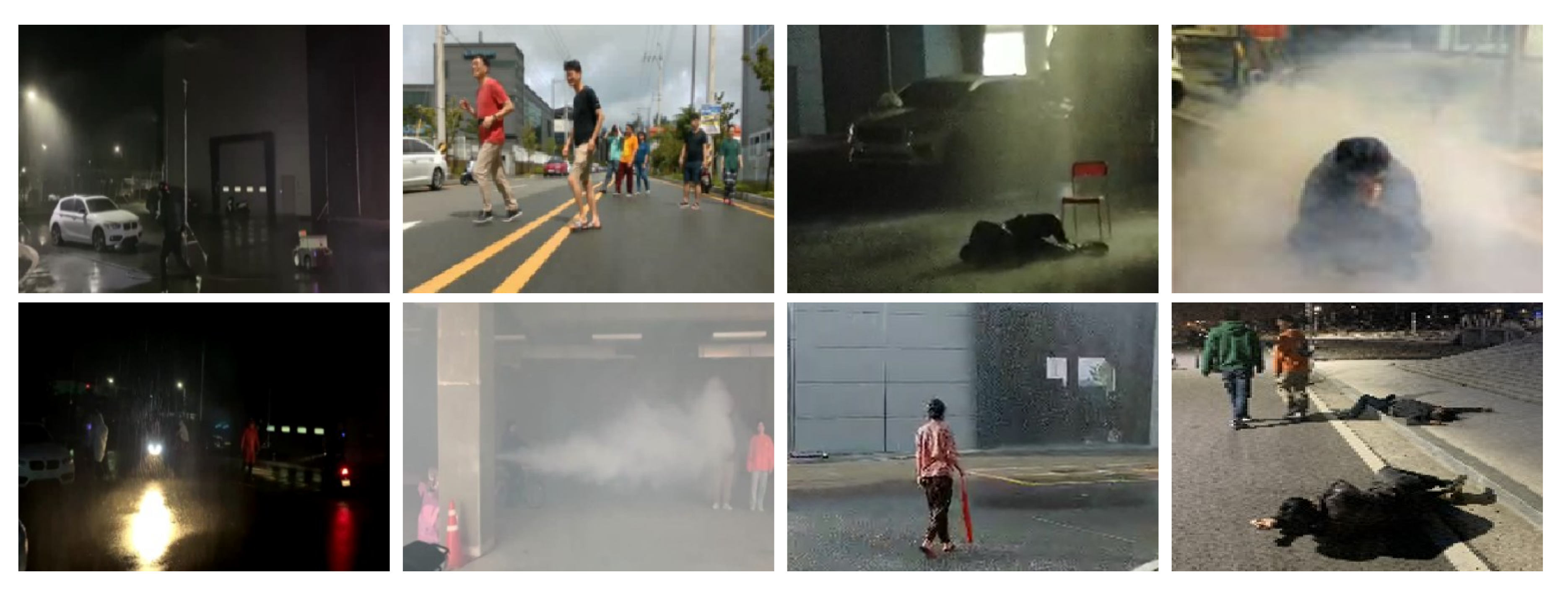}
\vspace{0.1cm}
\captionsetup{font=footnotesize}
\caption{Environmental diversity in our dataset: running at night, lying in fog, a walking person with an umbrella in the rain, a fallen person beside a chair in the heavy rain at night with various props (towel, umbrella, hard cap, etc.) and vehicles (car, bicycle, motorcycle, etc.). Our dataset also consists of various places (pavements, roads, warehouses, etc.).\label{fig_scenarios}}
\vspace{-0.2cm}
\end{figure} 
\vspace{-0.1cm}
Even though various deep-learning-based methods related to surveillance tasks using multi-modal sensors have been introduced~\cite{vertens2020heatnet,choi2016thermal,9664374}, there are still only a few multi-modal datasets for the study of surveillance tasks. For this reason, we have collected various requirements related to outdoor surveillance robots from researchers in a surveillance robot community and summarized them into three requirements: \textbf{a)} a dataset should be collected from a robot perspective view, i.e. first-person view data, \textbf{b)} a dataset should consist of multi-modal sensors, and \textbf{c)} a dataset should provide environmental diversity such as harsh weather and illuminance changes.

\begin{table*}[t!]
\vspace{0.25cm}
\centering
\captionsetup{font=footnotesize}
\captionsetup{justification=centering, labelsep=newline}
\caption{\sc{Dataset comparison with respect to sensor configuration and fitness for surveillance tasks}}
{\scriptsize
\begin{tabular}{c|ccccccc|c}
\toprule \midrule
\diaghead(3,-1){\hskip2.0cm}{Items}{\hspace{1.0cm}Datasets}  & \makecell{UCF\\Crime~\cite{sultani2018real}} &\makecell{3DPeS\\~\cite{baltieri20113dpes}}&\makecell{Monash \\Guns~\cite{lim2021deep}}&\makecell{2018 NVIDIA AI\\ City Challenge~\cite{naphade20182018}}&\makecell{Collective \\Activity~\cite{choi2009they}}&\makecell{ISR-UoL 3D\\Social Activity~\cite{coppola2016social}}&\makecell{NTU\\RGB+D 120~\cite{liu2019ntu}}&\thead{\textit{X-MAS}\\(proposed)}  \\ \midrule
 RGB & \checkmark&\checkmark  &\checkmark  &\checkmark  &\checkmark  &\checkmark  &\checkmark  &\checkmark  \\
 Depth &-&-  &-  &-  &-  &\checkmark  & \checkmark & \checkmark \\
 IR&-&-  &-  &-  &-  &-  &\checkmark & \checkmark \\
 Thermal&-&-  &-  &-  &-  &-  &-  & \checkmark \\
 Night Vision&-&-&-&-&-&-&-& \checkmark \\
 LiDAR&- &-  &-  &-  &-  &-  &-  & \checkmark \\ \midrule
 Scene label & \checkmark &\checkmark&\checkmark&\checkmark&\checkmark&\checkmark&\checkmark& \checkmark (each pair) \\
 Bounding box&-&\checkmark&\checkmark&\checkmark&\checkmark&- &- & \checkmark \\
 Mask image&-&-&-&-&-&-  &- & \checkmark \\ \midrule
 Tracking ID&-&\checkmark  &-  &-  &\checkmark  &-  &(Joint info.)& \makecell{\checkmark\\(only \textit{X-MAS}-C)} \\ \midrule
 Env./Scenario diversity&$\bigstar \bigstar \bigstar$&$\bigstar$&$\bigstar$&$\bigstar$&$\bigstar$&$\bigstar$&$\bigstar$&$\bigstar \bigstar \bigstar $\\ \midrule
 \makecell{Scale\\(V: Videos, I: Images)}&\makecell{>1.9K V\\(30 fps)}&200K I&2.5K I&\makecell{100 V (Track 2)\\(30 fps)}&\makecell{44 V\\($\approx$ 30fps)}&\makecell{80 V\\(30 fps)}&>8M I&>2.5M I\\
 \bottomrule \midrule
\end{tabular}
}
\label{ComparisionofDatasets}
\vspace{-0.9cm}
\end{table*}

\begin{figure}[t]
\centering
\includegraphics[width=8.5 cm]{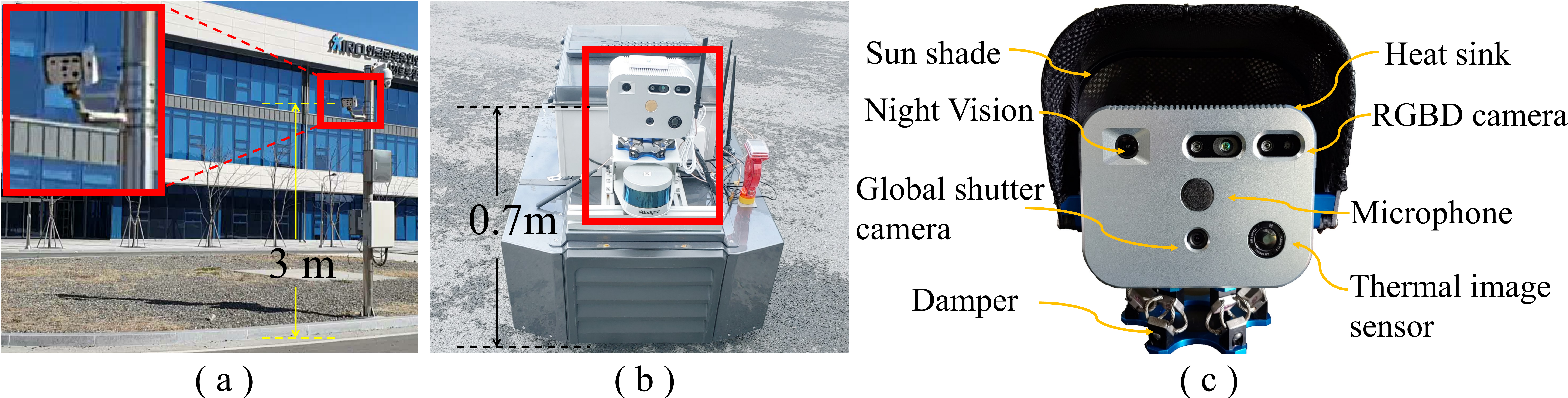}
\vspace{0.1cm}
\captionsetup{font=footnotesize}
\caption{Multi-agents~\cite{electronics11142214} used in this study. (a) Fixed sensor module. (b) A mobile robot equipped with the sensor module. Red boxes indicate sensor modules. (c) Configuration of the multi-modal sensor module. More detailed sensor specifications are provided in Appendix~\ref{sec:Sensor Specifications}.\label{fig_SensormodulesAndRobot}}
\vspace{-0.3cm}
\end{figure}

This study focuses on all three requirements for surveillance utilizing mobile robot platforms, as shown in Figs.~\ref{fig_example_of_dataset} and \ref{fig_scenarios}. In general, widely used public datasets~\cite{russakovsky2015imagenet,everingham2015pascal,lin2014microsoft} for a deep learning study mainly contain data in the view of image classification rather than a first-person view in terms of surveillance where the two types of the dataset are drastically different. For example, if the trained deep-learning based algorithm with third-person view datasets have been utilized for mobile robot platforms, this leads to performance degradation; therefore, some methods~\cite{baktashmotlagh2013unsupervised,aytar2011tabula} have been proposed for transforming third-person view datasets into first-person view. These methods still, however, have a limitation, i.e. they do not completely resolve the generalization issues yet. To tackle this limitation, we propose a robot perspective view dataset under real surveillance situations and environments using a mobile robot (see Fig.~\ref{fig_SensormodulesAndRobot}), and the dataset is named as \textit{X-MAS}, a combination of the words \textit{e\textbf{X}tremely large-scale, \textbf{M}ulti-mod\textbf{A}l, and \textbf{S}ensor}.

This study presents a dataset collected using multi-modal sensors including an RGB camera, a night vision, a thermal camera, and a depth camera in contrast to RGB-based datasets~\cite{sultani2018real,baltieri20113dpes,lim2021deep,naphade20182018,choi2009they}. The collected data reaches 2.5 million images. We also used professional actors to provide various scenarios and realistic actions for our dataset. The contributions of this paper are as follows: 

\begin{itemize}
\item We release a publicly available large-scale multi-modal dataset for the study of outdoor surveillance with high-quality annotations.   
\item In particular, the dataset includes time-synchronized and calibrated multi-mode sensor data that provide a first-person view gathered by a mobile robot.
\item Several use-cases related to utilizing this dataset are presented.
\end{itemize}

The remainder of this paper is organized as follows. Section~\ref{sec:related_work} presents related works, and Section~\ref{sec:multi_modal_data_collection} explains the system configuration, the dataset design,  and its contents. In addition, Section~\ref{sec:algorithmic_analysis} presents how to utilize the presented dataset and shows practical experiments. Finally, a summary of the findings and conclusions are presented in Section~\ref{sec:conclusion}.

\section{Related Works}\label{sec:related_work}
Many large datasets for object detection, tracking, segmentation, classification, and object recognition, have been released in the past decade. Popular RGB benchmark datasets such as ImageNet~\cite{russakovsky2015imagenet}, PASCAL VOC~\cite{everingham2015pascal}, and COCO~\cite{lin2014microsoft} can be utilized for these tasks, but are not affordable for harsh outdoor environments. 

\begin{table*}[t]
\vspace{0.25cm}
\captionsetup{font=footnotesize}
\captionsetup{justification=centering, labelsep=newline}
\caption{\sc{Our dataset design: e.g. a mobile robot captures a man (actor A) standing with a backpack on a rainy day in Pohang. }}
\begin{center}
{\scriptsize
\begin{tabular}{c|c|c|c|c|c|c|c|c}
\toprule \midrule
 Category & Agent & Place  & Situation   & Time   & Weather & Actor & Props & Scenario   \\ \midrule
 \makecell{Action classification} & \multirow{2}{*}{\makecell{Fixed-type/\\mobile agent}} & \multirow{2}{*}{\makecell{Pohang/\\Gwangju, \\ South Korea}} & \multirow{2}{*}{\makecell{Normal/\\anomaly}} & \multirow{2}{*}{\makecell{Day/\\night}} & \multirow{2}{*}{\makecell{Clear (night)/\\rainy/sunny}} & A/B/C/D & \makecell[l]{Hand-towels,\\backpack, etc.}  &\makecell[l]{Standing, convulsions, etc.} \\ \cline{1-1} \cline{7-9} 
 \makecell{Detection/tracking} &  &   &   &   &   &\makecell{Arbitrary\\(crowd)}&Arbitrary&\makecell[l]{Walking, fallen bicycle, etc.}\\ \midrule\bottomrule
\end{tabular}
}
\end{center}
\vspace{-0.9cm}
\label{ourDatasetDesign}
\end{table*}

Table~\ref{ComparisionofDatasets} summarizes existing public datasets for surveillance tasks compared to ours, and the details of them are introduced below. Although various sensor fusion algorithms have been introduced in the last decade, many studies have focused on RGB images and videos. Several RGB datasets are described as follows: The \textbf{UCF-Crime} dataset~\cite{sultani2018real}, which was collected from CCTVs, contains 1,900 videos of indoor and outdoor anomaly scenes, and day and night scenes. This dataset provides labels (normal or anomaly) for each video; however, it does not provide bounding boxes or masks for objects. Nevertheless, it has been widely used 
in learning-based abnormal behavior detection~\cite{zaheer2020self}. The \textbf{3DPeS} dataset~\cite{baltieri20113dpes} provides 500 videos for 3D/multi-view surveillance and forensic applications. It was designed to evaluate the performance of re-identification, segmentation, and detection/tracking of people. Since this dataset was collected by multiple cameras, it can be exploited to track and re-identify humans for surveillance. The \textbf{Monash Guns} dataset~\cite{lim2021deep} focuses on images of humans with a gun. It contains over 2,500 different CCTV images of guns in realistic settings.
\textbf{AI City Challenge}~\cite{naphade20182018} provides a new set of videos and images with different tasks every year.  Ritika  Bhardwaj \textit{et al.}~\cite{bhardwaj2021computationally} utilized this dataset for video traffic surveillance. The \textbf{Collective Activity} dataset~\cite{choi2009they} contains 44 short video sequences and five different collective activities of crossing, walking, waiting, speaking, and queueing, some of which were recorded with a consumer hand-held digital camera.

In addition, several datasets related to surveillance tasks were collected using an RGB-D sensor as follows: The \textbf{ISR-UoL 3D Social Activity} dataset~\cite{coppola2016social} from the Lincoln Center contains RGB, depth, and tracked skeleton data collected with RGB-D sensors. It includes eight social activities: handshake, greeting hug, help walk, help stand-up, fight, push, conversation, and call attention. Jun Liu \textit{et al.} presented the large-scale dataset \textbf{NTU RGB+D 120}~\cite{liu2019ntu} for RGB-D sensor based human action recognition, collected from 106 distinct subjects and containing more than 114,000 video samples and 8 million frames. 

In the past decade, many multi-modal datasets have been published for autonomous driving~\cite{9760091}. Di Feng \textit{et al}.~\cite{feng2020deep} introduced various multi-modal datasets, but most of them are  not  suitable for outdoor surveillance robots. Therefore, we present more than 500,000 annotated outdoor image pairs with synchronized and calibrated multi-modal sensor data, which were collected at a frame rate of 10 Hz for surveillance. In addition, we focused on the human activities with various props under surveillance situations. We suggest that the proposed dataset is useful for researching robust sensor fusion algorithms regarding outdoor surveillance tasks.
\vspace{-0.15cm}

\begin{figure}[b]
\vspace{0.1cm}
\centering
\includegraphics[width=6.0 cm]{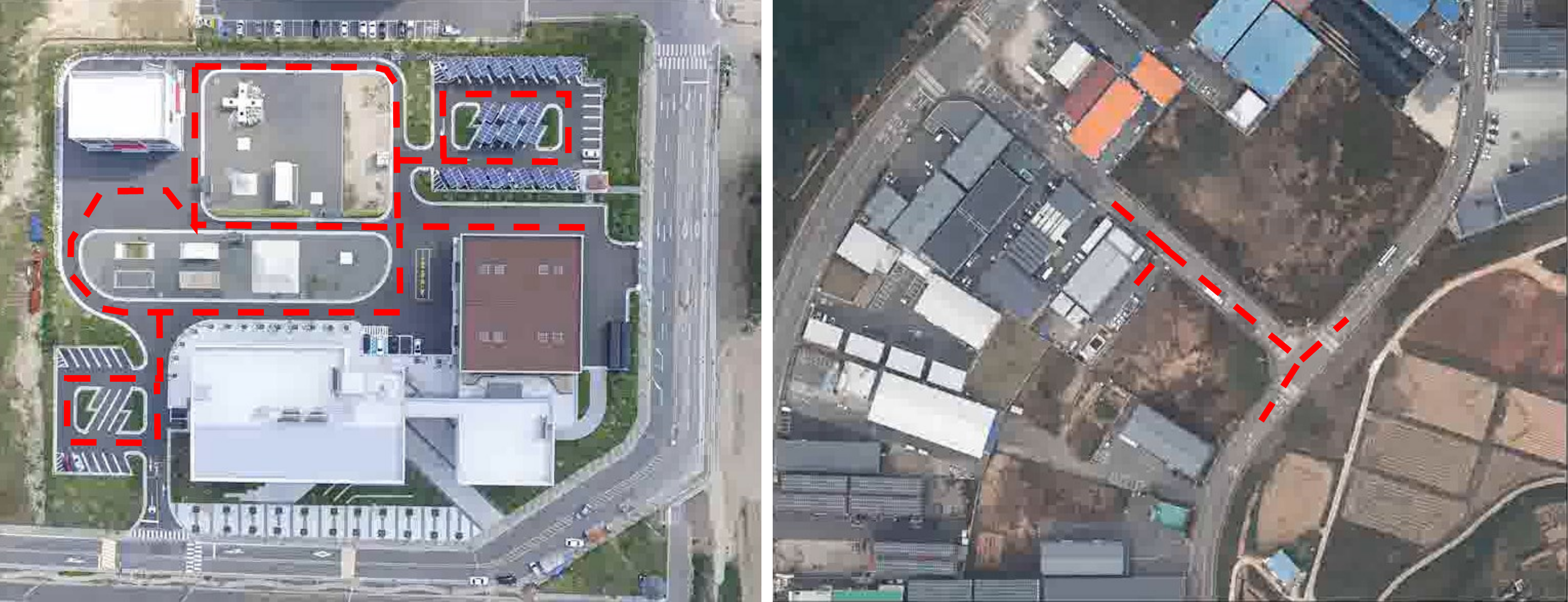}
\vspace{0.1cm}
\captionsetup{font=footnotesize}
\caption{Test beds: (left) the disaster robotics center in Pohang, and (right) the Nano industrial complex in Gwangju. Red dashed lines indicate robot paths.\label{fig_testbed}}
\end{figure}

\section{System Configuration and Multi-modal data collection}\label{sec:multi_modal_data_collection}
\vspace{-0.15cm}
The dataset was collected using a cloud-based surveillance system at Pohang and Gwangju in South Korea between April 2017 and August 2021. The system was composed of mobile robots, fixed sensor modules, and a cloud system for outdoor surveillance. The mobile robot was equipped with a multi-modal sensor module, as depicted in Fig.~\ref{fig_SensormodulesAndRobot}. It patrolled around the building and road at a speed of approximately 0.4m/s, as shown in Fig.~\ref{fig_testbed}. All data were annotated by well-trained annotators, and our dataset provides the annotation information as XML files.    

\vspace{-0.15cm}
\subsection{Environments}
\vspace{-0.1cm}

A robot patrols around the building, and fixed cameras are installed at the borders of the building and significant points to collect the multi-modal sensor dataset for surveillance tasks. Therefore, we selected two big sites for operating our surveillance robots, as shown in Fig.~\ref{fig_testbed}. The robot moved along the pavement around the buildings under various weather conditions including rainy and foggy days. Walker, cyclists, and cars pass by the mobile robots and fixed multi-modal sensor module. While the mobile robot patrols, humans, cyclists, and cars are observed. 


\begin{figure}[b]
\vspace{-0.7cm}
\centering
\includegraphics[width=9.0 cm]{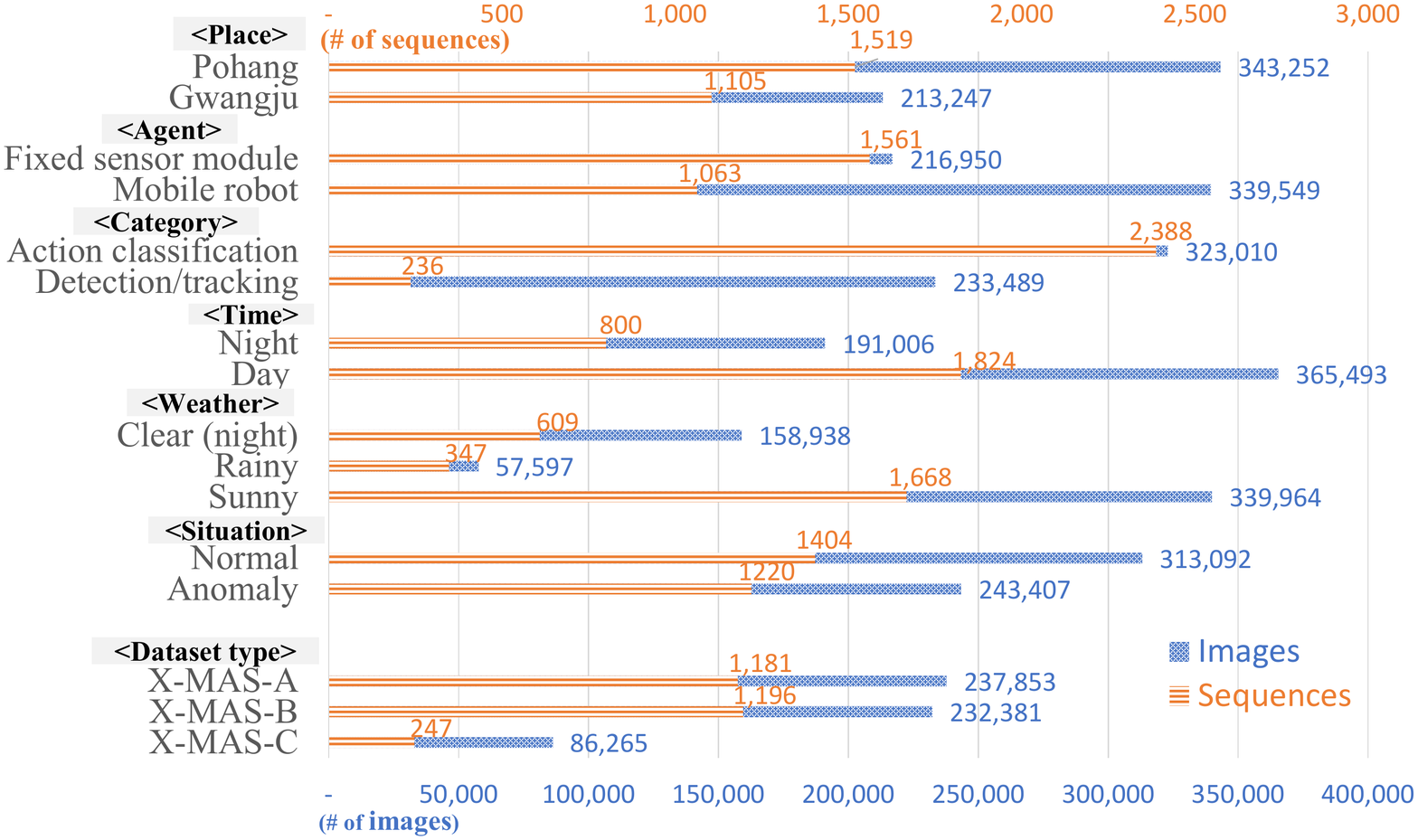}
\captionsetup{font=footnotesize}
\vspace{-1.0cm}
\caption{Statistics of the dataset: The orange-colored bar indicates the number of sequences, and the blue-colored bar indicates the number of image pairs.}
\label{fig_statisticsofdataset_total}
\end{figure} 

\vspace{-0.2cm}
\subsection{Multi-Modal Sensor Module}
\vspace{-0.15cm}

This study presents a sensor module with complementary physical sensors. Robots must utilize more than one sensor to perform an outdoor surveillance task around the clock. Therefore, our sensor module includes five different sensors, as shown in Fig.~\ref{fig_SensormodulesAndRobot} (c). Six types of calibrated dataset are obtained using this sensor module. In particular, the sensor module contains a damper that is capable of mitigating vibrations of different frequencies, allowing for clear images to be collected while the mobile robot is moving. Furthermore, we synchronized all the sensors within a reasonable time interval. For the detailed information on the sensor configuration, calibration, and synchronization, please refer to our previous work~\cite{electronics11142214}. 

\begin{figure*}[t!]
\centering
\includegraphics[width=18cm]{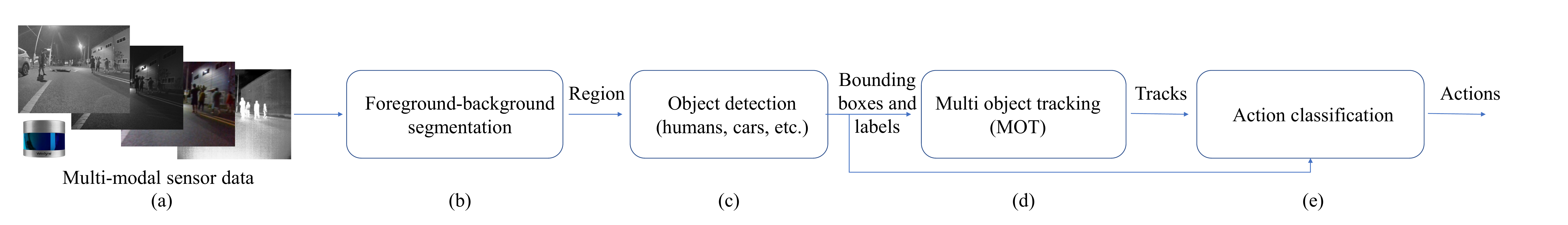}
\vspace{0.25cm}
\captionsetup{font=footnotesize}
\vspace{-0.3cm}
\caption{Pipeline of the surveillance system: (a) multi-modal sensor data collection, (b) foreground separation from an image to reduce search space, (c) object detection regarding objects that is relevant to surveillance, (d) multi-object tracking for intruder tracking using data association with range data, (e) action classification regarding a primitive action (stand, lie, sit) of each detected person in each scene. Outputs of each module ((c), (d), (e)) are exploited for autonomous driving of robots and accumulated in a scene understanding module that analyzes a scene whether it is normal or anomaly (See GitHub ({https://github.com/lge-robot-navi})). In addition, our study focuses on real-time performance (> 10 fps) for the surveillance purpose.}
\label{fig_framework_wholeSystem}
\vspace{-0.7cm}
\end{figure*} 

\begin{table}[b]
\captionsetup{font=footnotesize}
\captionsetup{justification=centering, labelsep=newline}
\caption{\sc{Scenarios for the action classification}}
\begin{center}
	\setlength{\tabcolsep}{4pt}
{\scriptsize
\begin{tabular}{c|l|c}
\toprule \midrule
 No. & Scenario (see images in the GitHub repository)  &  Situation    \\ \midrule
 1 &  Lying down on the ground & \multirow{8}{*}{Normal}\\
 2 &  Standing &\\
 3 &  Walking &\\
 4 &  Running &\\
 5 &  Sitting on a chair &\\
 6 &  Sitting on the ground &\\
 7  &  Standing $\rightarrow$ sitting on the ground $\rightarrow$ standing &\\
 8 &  Standing $\rightarrow$ sitting on a chair $\rightarrow$ standing &\\
 \midrule
 9 &  Sitting on a chair $\rightarrow$ falling down & \multirow{8}{*}{Anomaly}\\
 10 &  Sitting on the ground $\rightarrow$ falling down &\\
 11 &  Standing $\rightarrow$ falling down &\\
 12 &  Walking $\rightarrow$ falling down &\\
 13 &  Running $\rightarrow$ falling down &\\
 14 &  Convulsions &\\
 15 &  A drunken human &\\
 16 & Etc. &\\
\bottomrule \midrule
\end{tabular}
}
\end{center}
\label{scenario_actionDB}
\end{table}

\vspace{-0.2cm}
\subsection{Dataset Design} 
\vspace{-0.15cm}

Recent studies have focused on the context-awareness of multiple scenes with sensor fusion and deep learning algorithms~\cite{https://doi.org/10.4218/etrij.2021-0395}. Thus, our dataset provides sequential images (more than 100) for each episode named as a sequence. These multiple episodes are divided into normal or anomalous situations. In addition, our dataset has high diversity in the surveillance domain, including various places, humans with various props, and different weather conditions as summarized in Table~\ref{ourDatasetDesign}. Classifying the human actions is equally important as human tracking because most surveillance situations, in general cases, are related to the actions of humans. Thus, we separated the dataset into two categories: \emph{action classification} and \emph{detection/tracking}.

\subsection{Data Statistics}
In summary, the \textit{X-MAS} dataset contains 2,624 sequences and 556,499 pairs and is analyzed by place, agent, category, time, weather, and situation, as shown in Fig.~\ref{fig_statisticsofdataset_total}. The number of action classification sequences is 2,388 and the number of detection/tracking sequences is 236.

\subsection{Data Annotation}

\begin{table}[t]
\captionsetup{font=footnotesize}
\captionsetup{justification=centering, labelsep=newline}
\caption{\sc{Scenarios for detection/tracking}}
\begin{center}
	\setlength{\tabcolsep}{4pt}
{\scriptsize
\begin{tabular}{c|l|c}
\toprule \midrule
 No. & Scenario (see examples in the GitHub repository) &  Situation \\ \midrule
 1 &  Walking& \multirow{4}{*}{Normal}\\ \cline{2-2}
 2 &  Walking + running &\\ \cline{2-2}
 3 &  Walking beside a car and a motorcycle &\\ \cline{2-2}
 4 &  Walking + running beside a car and a motorcycle &\\
 \midrule
 5 &  Walking + fallen or sat people & \multirow{6}{*}{Anomaly}\\ \cline{2-2}
 6 &  Walking + running + fallen or sat people &\\ \cline{2-2}
 \multirow{2}{*}{7}&  Walking + running +  &                   \\
   &  fallen or sat people beside a car and a motorcycle & \\ \cline{2-2}
 8 &  Walking + running beside a fallen bicycle &\\ \cline{2-2}
 9 &  Walking + running beside a fallen motorcycle &\\
\bottomrule \midrule
\end{tabular}
}
\end{center}
\vspace{-0.4cm}
\label{scenario_detectionDB}
\end{table}

For this reason, the annotations of our \textit{X-MAS} are mainly categorized into two parts. First, the action classification dataset provides XML files and mask images as shown in Fig.~\ref{fig_example_of_dataset} (e). The XML files describe the condition, scenario, and the name of a mask image (visit our GitHub ({https://github.com/lge-robot-navi})). 
The dataset for detection and tracking provides only XML files without mask images. It provides bounding boxes that depict the locations and classes of objects for both RGB and thermal images, thus enabling detection algorithms to perform detection on both modalities. In addition, the dataset has been collected over five years, and there are several types of datasets, for example; a dataset with/without tracking scenarios and a dataset with/without tracking IDs. The category of each dataset is indicated by sub-folder names (see Appendix~\ref{sec:Characteristics of Each Dataset}).

\subsection{Description of Various Scenarios}
In this study, professional actors acted with various props, such as backpacks and hard hats, for collecting various appearance and situation dataset. This work was conducted with predefined scenarios under several weather conditions, and each scenario consisted of more than 100 sequential images. Additionally, pixel-level masking data for object segmentation is provided. The scenarios for action classification are summarized in Table~\ref{scenario_actionDB}.

The scenarios for detection and tracking are summarized in Table~\ref{scenario_detectionDB}. The dataset for detection and tracking was collected by professional actors and researchers under various weather conditions. Each scenario consisted of at least 300 sequential images.

\section{Use Cases of Developing Surveillance Algorithms using the dataset}\label{sec:algorithmic_analysis}
This section provides several ways to utilize two categories of the presented dataset for outdoor surveillance studies. The action-classification dataset can be mainly exploited for foreground-background segmentation and action/scene classification. On the other hand, the detection/tracking dataset can be exploited for studies of surveillance tasks that contain foreground-background segmentation, object detection, tracking, and action classification (see Fig.~\ref{fig_framework_wholeSystem}). Note that we open our source code used in our surveillance system via GitHub\footnote[1]{https://github.com/kyuewang17/SNU\_USR\_dev}. In the following sub-sections, the study presents various approaches for each module and results.      

\vspace{-0.15cm}
\subsection{Foreground-Background Segmentation}
\begin{figure}[t]
\centering
\includegraphics[width=8.5cm]{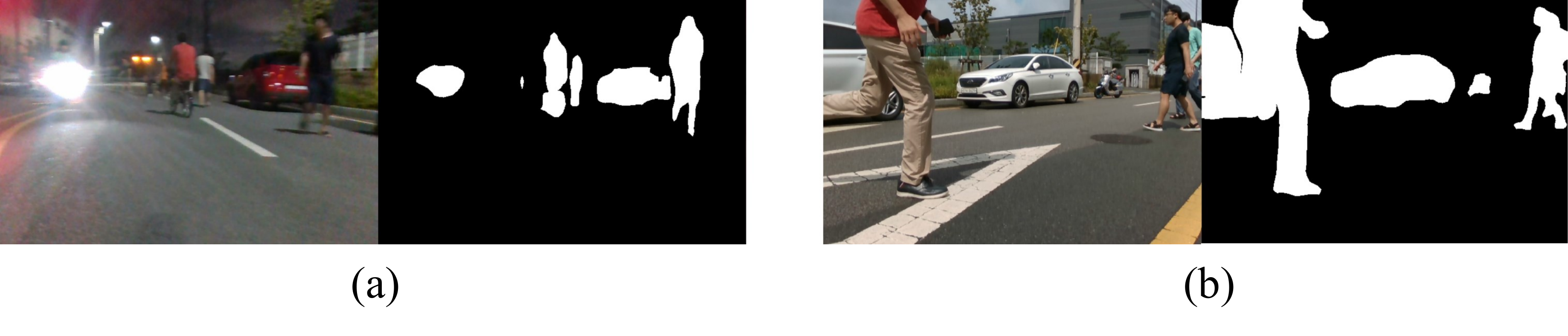}
\captionsetup{font=footnotesize}
\caption{The qualitative results of DeepLabv3+~\cite{chen2018encoder} from samples of detection/tracking dataset during (a) the nighttime RGB sequence (1-10d) and (b) the daytime RGB sequence (1-07d) at the same place.}
\label{fig_segmentation}
\vspace{-0.3cm}
\end{figure}

The action classification dataset provides pixel-level masks for evaluating foreground-background segmentation, and it can be exploited for the quantitative performance evaluation. The detection/tracking dataset also can be exploited for the qualitative performance evaluation, as shown in Fig.~\ref{fig_segmentation}. Even though the detection/tracking dataset provides no pixel-level masks, it contains various dynamic objects and environments; therefore, it is also suitable for foreground-background segmentation research on mobile robot surveillance. We leveraged four kinds of segmentation networks for separating the foreground and evaluated the quantitative performance on the CDNet dataset~\cite{wang2014cdnet} as summarized in Table~\ref{Performance_comparison_seg}. The trend between the quantitative result on the CDNet dataset and the qualitative results utilizing the \textit{X-MAS} dataset is experimentally equivalent. Our findings are as follows: a) a feature pooling module is better than a multi-scale encoder in the view of performance, b) depth-wise separable convolution with ResNet-101 as a backbone is more efficient regarding the F-score, being applied to DeepLabv3+~\cite{chen2018encoder}, and c) ResNet-101 is more appropriate than VGG-16 and ResNet-34 in our surveillance application. Based on it, we leveraged DeepLabv3+~\cite{chen2018encoder} based segmentation network. In addition, the COCO~\cite{lin2014microsoft} dataset was used for pre-training the network to separate objects more efficiently in an outdoor surveillance scenario. 
\vspace{-0.15cm}
\subsection{Object Detection}
The goal of this paper is to offer a dataset that enables the users make a network in various harsh environments such as rainy, foggy, and night. Thus, the dataset provides rich data in various harsh environments as shown in Fig.~\ref{fig_scenarios}.
We have employed several state-of-the-art models such as Faster R-CNN, Cascade R-CNN, Deformable
DETR, DETR, and YOLOv5\footnote[2]{https://github.com/ultralytics/yolov5}. YOLOv5 model was superior to other models in terms of performance and real-time processing in this task and was applied to our surveillance robots. 

\begin{table}[t]
\vspace{0.25cm}
\captionsetup{font=footnotesize}
\captionsetup{justification=centering, labelsep=newline}
\caption{\sc{Performance comparison on the CDNet dataset~\cite{wang2014cdnet} between our approaches for foreground-background segmentation.}}
    \begin{center}
    {\scriptsize
    \begin{tabular}{l|c|cc}
      \toprule \midrule
        \multirow{2}{*}{Methods} & \multirow{2}{*}{Backbone}  & \multicolumn{2}{c}{F-score}     \\ \cline{3-4}
          &           & \multicolumn{1}{c|}{Day} &    Night                       \\ \midrule
          Auto-encoder w/ multi-scaled images& \scriptsize{VGG-16} & \multicolumn{1}{c|}{\scriptsize{0.92}} &    \scriptsize{0.68}               \\ 
          FgSegNet V2 w/ feature pooling     & \scriptsize{ResNet-34}         & \multicolumn{1}{c|}{0.94} &    0.71           \\
        \midrule
        \multicolumn{1}{l|}{\makecell[l]{FgSegNet V2 w/ Atrous and \\ depth-wise separable convolution}} & \scriptsize{ResNet-34} & \multicolumn{1}{c|}{\scriptsize{0.94}} & \multicolumn{1}{c}{\scriptsize{0.72}}             \\ 
        \midrule
        \multicolumn{1}{l|}{DeepLabv3+~\cite{chen2018encoder}} & \scriptsize{ResNet-101}& \multicolumn{1}{c|}{\scriptsize{0.99}} & \multicolumn{1}{c}{\scriptsize{0.92}}             \\ 
        \bottomrule \midrule
        \end{tabular}
        }
    \end{center}
\label{Performance_comparison_seg}
	\vspace{0.1cm}
\end{table}

\begin{table}[t]
\vspace{0.25cm}
\captionsetup{font=footnotesize}
\captionsetup{justification=centering, labelsep=newline}
\caption{\sc{Performance comparison between models pre-trained only with the COCO dataset only and models fine-trained with our multi-modal sensor dataset.}}
    \begin{center}
    {\scriptsize
        \begin{tabular}{l|cccc}
        \toprule \midrule
        \multirow{2}{*}{Models} & \multirow{2}{*}{\makecell{COCO only\\(mAP, \%)}} & \multicolumn{3}{l}{Ours (\textit{X-MAS} + COCO), (mAP, \%)} \\ \cline{3-5}
        
                  &                   &  Case 1    & Case 2  & Case 3      \\ \midrule
             Faster R-CNN & 5.7 & 45.7 & 31.9& 12.7 \\
             Deformable DETR & 5.5 & 51.0 &32.3 & 14.0\\
             Cascade R-CNN & 5.9 & 51.1 &34.6 &25.5 \\
             DETR & 6.1 & 57.0 & 33.3&20.4 \\
        \midrule
             YOLOv5 & 5.5 & 58.7 &32.5&19.6\\
        \bottomrule \midrule
        \end{tabular}
        }
        
        \raggedright \scriptsize{Case 1: RGB + Thermal + COCO, Case 2: RGB + COCO, Case 3: Thermal + COCO}
    \end{center}

\label{quant_od}
	\vspace{0.1cm}
\end{table}

\begin{figure}[t]
\centering
\includegraphics[width=5.0cm]{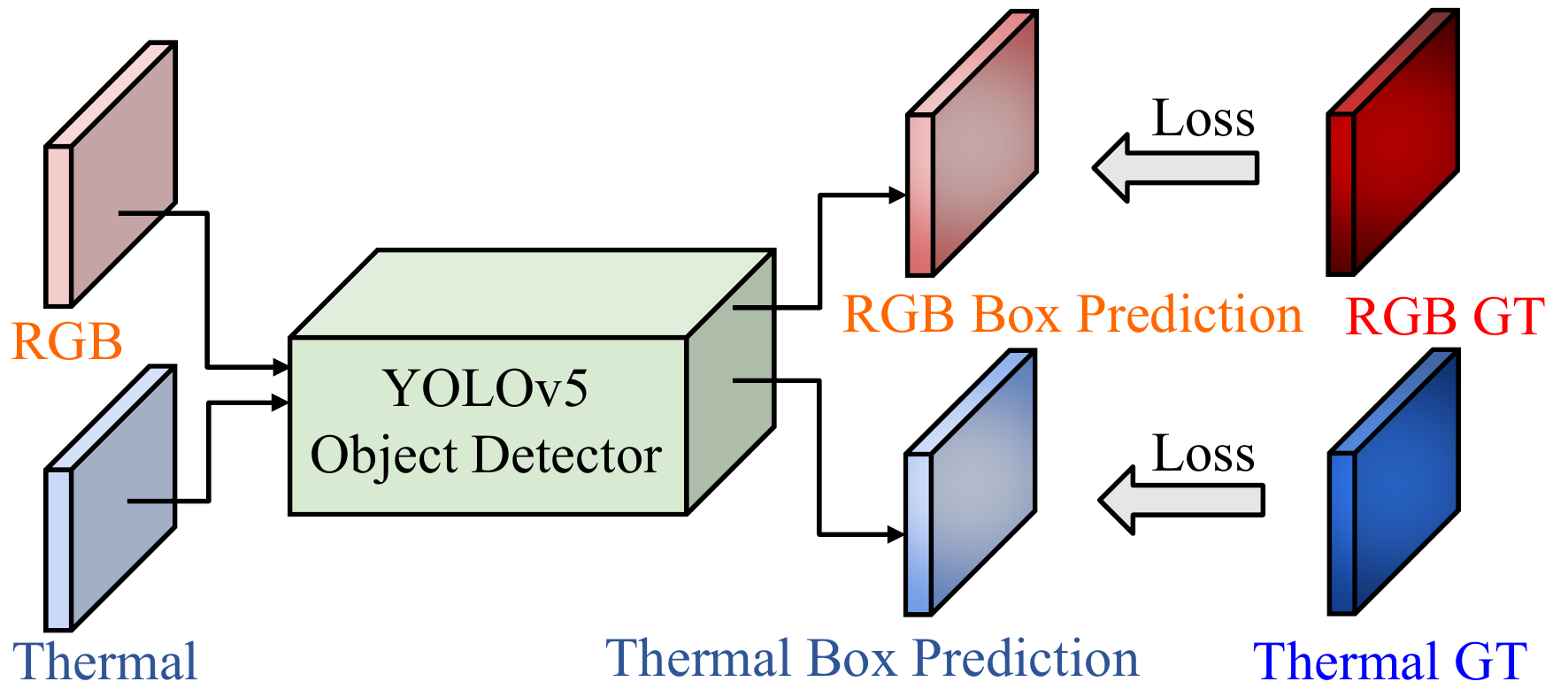}
\vspace{0.1cm}
\captionsetup{font=footnotesize}
\caption{Framework of the object detector training using our multi-modal sensor dataset.}
\label{fig_frameworkofobjectdetector}
	\vspace{-0.2cm}
\end{figure}

\begin{figure}[b]
\vspace{-0.1cm}
\centering
\includegraphics[width=7.0cm]{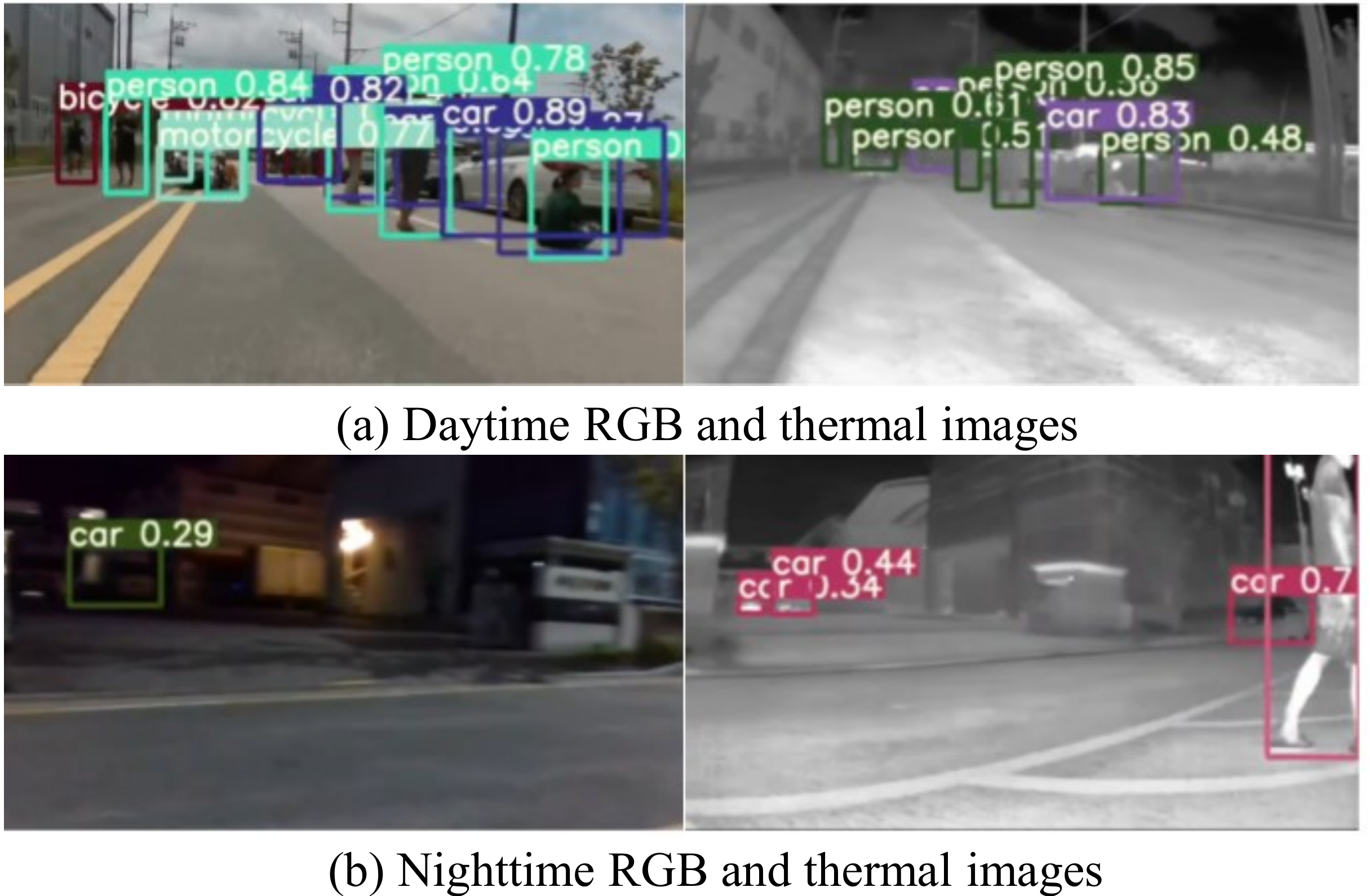}
\vspace{0.1cm}
\captionsetup{font=footnotesize}
\caption{Qualitative results of the object detector trained with our multi-modal dataset. (a) shows its prediction results on a daytime sequence (1-07d), and (b) shows that of a nighttime sequence (1-10d).}
\label{qual_od}
\end{figure} 

For the training, we used our own multi-modal sensor dataset and the COCO~\cite{lin2014microsoft} training set. Specifically, we split our multi-modal dataset into training and validation sets. Images were sampled from several sequences (see Appendix~\ref{sec:Reduced Dataset}) of our dataset for the evaluation and utilized the rest of the data for the training. 
Daytime and nighttime images are labeled on RGB and thermal images respectively, hence the detector was trained with both RGB and thermal images, as shown in Fig.~\ref{fig_frameworkofobjectdetector}.

Table~\ref{quant_od} shows the performance comparison between several models. We trained several models with different training data settings to demonstrate the usefulness of our dataset. In Table~\ref{quant_od}, ``COCO only'' column presents the performances of models trained with the COCO train set only, and ``Ours (\textit{X-MAS} + COCO)'' column presents the performances of models trained not only with the COCO training set but also with our dataset.  All the models show better performance for both RGB and thermal images in the case that we additionally used our multi-modal dataset during the training phase, as listed in Table~\ref{quant_od}.

Fig.~\ref{qual_od} shows the qualitative results of the object detector trained with our multi-modal dataset. As shown in Fig.~\ref{qual_od}, the prediction results on RGB images are better in the daytime, whereas the prediction results on thermal images are better in the nighttime.
The prediction results of the two modalities complement each other, thereby enabling the object detector to be robust in both day and night time. Through both quantitative and qualitative results, we demonstrate that our dataset can be used to train and evaluate the detector for both RGB and thermal images to robustly detect objects of interest during day and night time.

\subsection{Multi-Object Tracking}

In the field of computer vision, object tracking research has been separately studied as a study to track a single-object or a study to track multiple objects. The notable difference between single-object tracking and multi-object tracking (MOT) task is the initialization of tracks. 

One of the most widely known datasets for \textit{visual tracking} is the OTB-100\cite{wu2013online}. It contains different classes, sizes, and aspect ratios in various situations, including background clutter, illumination variation, fast motion, and target deformation. However, most of the data in the OTB-100 dataset and more recent datasets\cite{fan2019lasot,muller2018trackingnet,huang2019got} for tracking are less relevant to unmanned surveillance situations because the datasets were collected from sports, animal, and everyday videos. On the other hand, our detection/tracking dataset is suitable for multi-object tracking tasks in surveillance situations. In particular, the \textit{X-MAS-C} dataset provides tracking IDs for evaluating quantitative performance on diverse algorithms. 

\begin{figure}[b]
\centering
\vspace{-0.2cm}
\includegraphics[width=7.5cm]{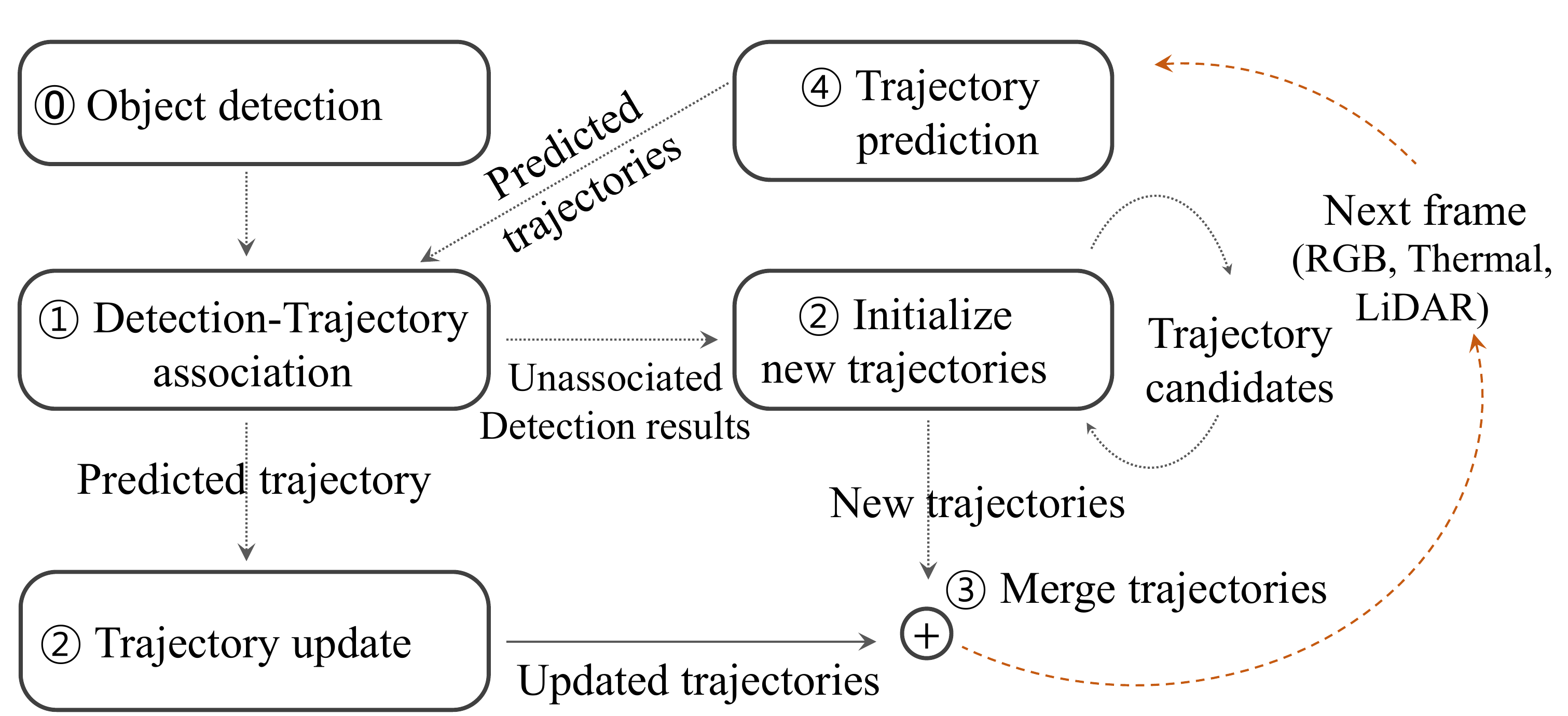}
\captionsetup{font=footnotesize}
\caption{Pipeline of the proposed tracker using a 3D Kalman filter}
\label{fig__multi_modal_mot_pipeline}
\end{figure}

\begin{figure}[t]
\centering
\includegraphics[width=6.5cm]{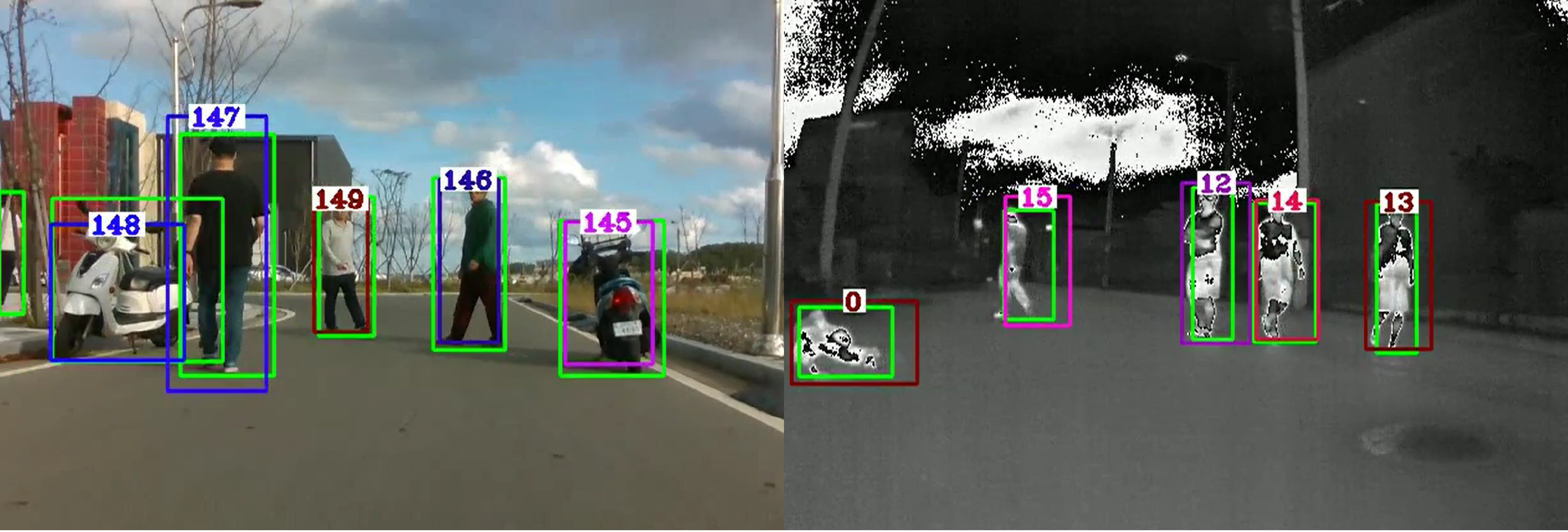}
\vspace{0.1cm}
\captionsetup{font=footnotesize}
\caption{Qualitative results of the multi-modal multi-object tracker. Left is the result image for daytime~(RGB). Right is the result image for nighttime~(thermal). Green and the other colors indicate the ground truth and tracking results, respectively.}
\label{fig__qualitative_results_mot}
\vspace{-0.4cm}
\end{figure}

Real-time operability and reliability are essential requirements when developing multi-object trackers for surveillance robots. To meet real-time operability, we excluded many recent deep learning-based methods that require computational complexity and referred to the source code of SORT~\cite{7533003}. Whereas SORT is a 2D MOT algorithm for RGB videos, our MOT algorithm is designed for multi-modal image sensor input. Therefore, using range and thermal data in our dataset, such as depth images and LiDAR data, a tracking-by-detection method that utilizes 3D (a bounding box center position (2D) and an actual depth value (1D)) Kalman filtering framework\cite{kalman1960new} and the Hungarian association algorithm\cite{kuhn1955hungarian} is proposed, as illustrated in Fig.~\ref{fig__multi_modal_mot_pipeline}. As summarized in Table~\ref{table__mot_results_on_mmosdx-c_dataset}, the study presents a performance baseline on the dataset sampled from the \textit{X-MAS-C}, and Fig.~\ref{fig__qualitative_results_mot} shows the qualitative results of the proposed tracker during daytime and nighttime.  The algorithm is conducted in real-time (>30Hz) in our system (Jetson AGX Xavier). The dataset can be exploited for studying various multi-object trackers, including our approach. 

The tracking process is briefly described as follows. In each multi-modal sensor frame, detection results are obtained from the object detector~(\,\textbf{Step 0}\,). In succession, the detection results are associated with the trajectory of objects~(\,\textbf{Step 1}\,), and their states in the current frame are predicted with the Kalman state transition matrix~(\,\textbf{Step 4}\,). The association is conducted via the Hungarian algorithm, and the association cost matrix is computed based on the intersection-over-union~(IOU), center distance, and histogram cosine similarity between the patches from the detection results and trajectory of objects. The patches are achieved from RGB images in daytime and thermal images in nighttime. The daytime and nighttime are classified via sunrise/sunset clock time. \textbf{Step 2} includes two processes: \textit{trajectory initialization} and \textit{trajectory update}. In the trajectory initialization step, new trajectories of objects are generated from detection results that fails to associate in \textbf{Step 1}. Meanwhile, the states of the trajectories of objects are updated via the Kalman update process in the trajectory update step. Then, the new and updated trajectories of objects are merged, and the memory for these objects is kept until the next frame for trajectory prediction step~(\,\textbf{Step 4}\,), recursively. For the trajectory state which is updated and predicted via the Kalman filter, the state vector includes the trajectory's center position, velocity, size, and depth information~(i.e. 3D Kalman filter). The trajectory depth value is computed via LiDAR point-cloud data, between \textbf{Steps 1} and \textbf{2} in every frame. The LiDAR sensor is calibrated with both RGB and thermal cameras; thus we can utilize projected~(2D) point-clouds regardless of time. In addition, we only project point-clouds that would be projected to the interior of the trajectories' bounding box for computational efficiency. The trajectory depth value is computed using the weighted sum of all 2D point-clouds inside the trajectory's bounding box. The weight is designed to be higher if 2D point-cloud data are closer to the bounding box center than the other 2D point-clouds, with a total sum of 1. 
\vspace{-0.25cm}


\begin{table}[t]
\vspace{0.25cm}
\captionsetup{font=footnotesize}
\captionsetup{justification=centering, labelsep=newline}
\caption{\sc{Quantitative results on our \textit{X-MAS}-C dataset that contains the ID annotations of objects for
tracker evaluation. }} 
\vspace{0.2cm}
\centering
	\setlength{\tabcolsep}{4pt}
 {\scriptsize
    \begin{tabular}{c|c|ccc} 
    \toprule \midrule
    Time & \textit{X-MAS}-C & MOTA$^3$ (\%) & Precision$^4$ (\%) & Recall$^5$ (\%) \\ \midrule
    Day (RGB) & MV1 & 38.5 & 65.3 & 94.5 \\ \midrule
    \multirow{2}{*}{\begin{tabular}[c]{@{}c@{}}Night\\ (Thermal)\end{tabular}} & 1-05d & 62.2 & 95.5 & 67.2\\ \cline{2-5}
    & 1-10d & 66.8 & 94.4 & 73.9 \\ \bottomrule \midrule
    \end{tabular}
    
    \raggedright \scriptsize{$^{3}$MOTA (Multiple Object Tracking Accuracy), $^4$precision, and $^5$recall metrics follow the definition from the MOT16 challenge~\cite{milan2016mot16}. Particularly, ``MOTA'' indicates the overall multi-object tracking performance in the given video sequence.}

    }
\label{table__mot_results_on_mmosdx-c_dataset}
\vspace{0.5cm}
\end{table}

\subsection{Action Classification}
\vspace{-0.15cm}
\begin{figure}[b]
\centering
\includegraphics[width=7.5cm]{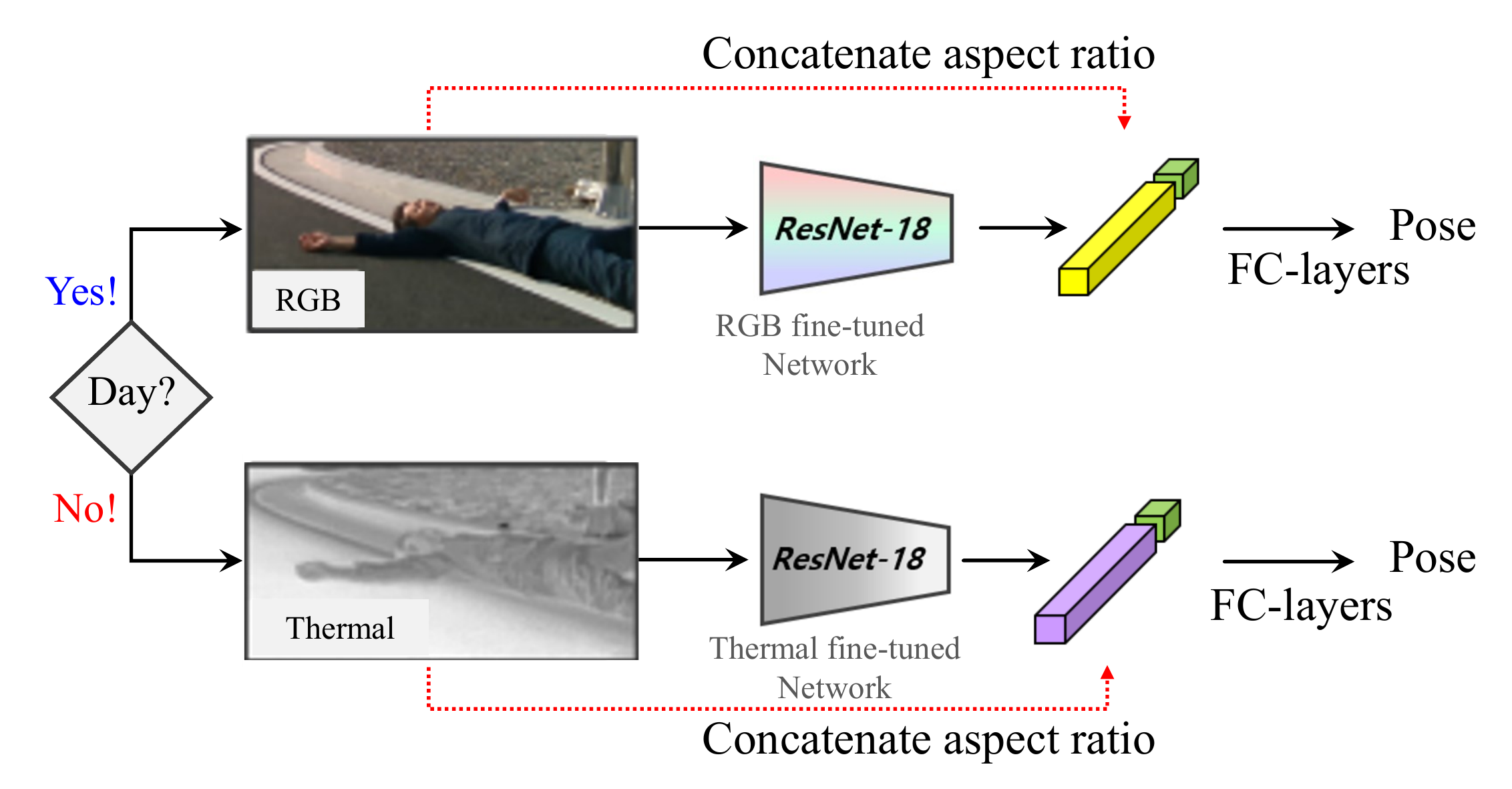}
\captionsetup{font=footnotesize}
\caption{Framework of the action classifier for real-time primitive actions.
\label{fig_Frameworkofthesimple_action_classifier}}
\end{figure} 

\begin{figure}[t]
\vspace{0.25cm}
\includegraphics[width=8.5cm]{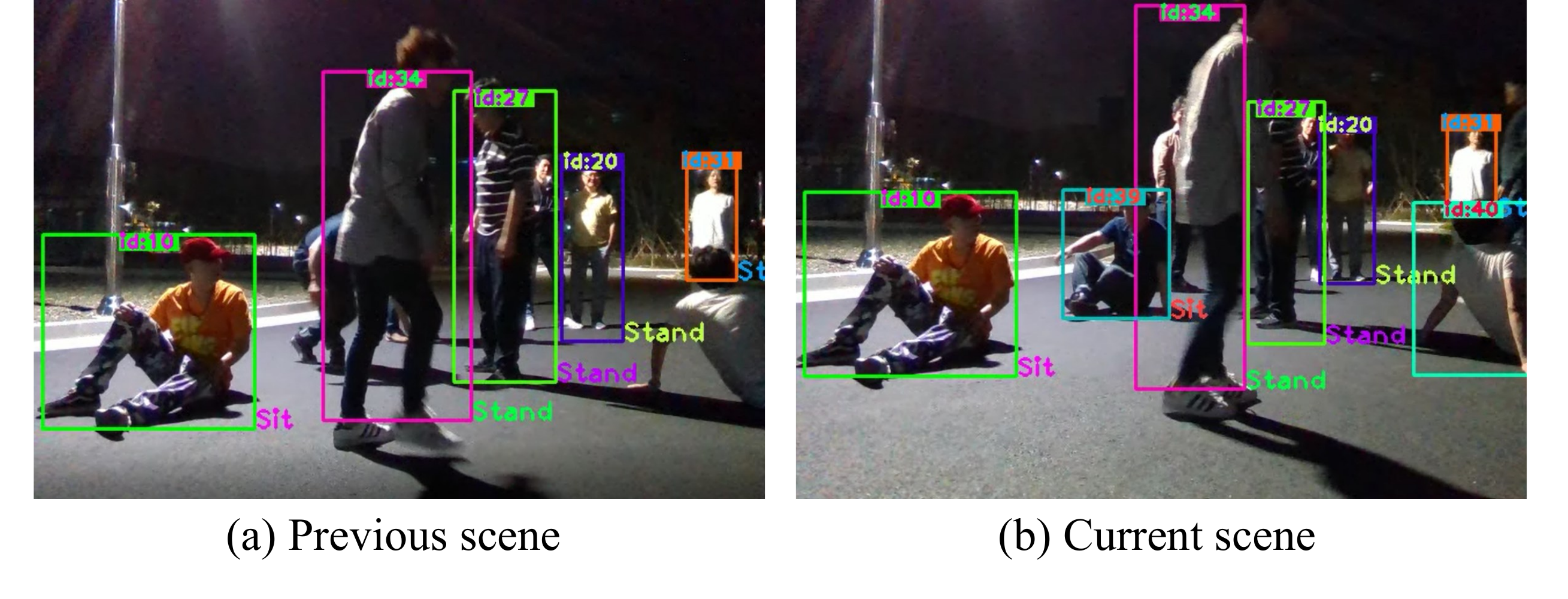}
\vspace{0.1cm}
\captionsetup{font=footnotesize}
\vspace{-0.2cm}
\caption{Qualitative results of the real-time action classifier and the object tracker for a mobile surveillance robot using sequential images. \label{fig_resultoftheactionclassifier}}
\end{figure} 

This subsection presents the use-case related to action classification using the dataset. Regarding action classification, the UCF~\cite{soomro2012ucf101} dataset has been widely used for developing and evaluating action classification algorithms; however, it is not directly related to the surveillance task. Action classification and situation awareness algorithms related to anomaly detection can be performed using sequential images of our dataset, and performance evaluation can also be conducted using the provided ground truth. Moreover, this dataset can be used in two ways: First, this dataset can be used to develop a one-class classification and a binary classification regarding anomaly detection using the action classification dataset, as described in Table~\ref{scenario_actionDB}.  Second, using the dataset, we can evaluate action recognition algorithms for primitive actions (standing, sitting, and lying) scene by scene. 

The study demonstrates real-time primitive action classification and evaluates the performance using annotations of each person's actions in frames. We modified ResNet-18~\cite{he2016deep} such that an aspect ratio can be used by concatenating an aspect ratio to a feature vector output from the convolutional layers of ResNet-18. Moreover, the concatenated features are fed to the following FC layers of ResNet-18. Our action classifier consists of each classifier for RGB and thermal image inputs. Switching between the two classifiers depending on whether it is daytime or not enables robust classification, as presented in Fig.~\ref{fig_Frameworkofthesimple_action_classifier}.

For the training, we sampled human patches from sequences 1-03 (day) and 1-09 (night) of our action classification datasets. We split sampled human patches into training and test sets for each sequence. We trained a classifier for RGB images on the 1-03 training set (day) and the classification accuracy was $95.6\%$ on the test set. For a thermal images, we trained a model on the 1-09 training set (night) and the classification accuracy was $93.4\%$ on the test set. Additionally, Fig.~\ref{fig_resultoftheactionclassifier} shows the qualitative results of action classifications for each tracked person.

\vspace{-0.1cm}
\section{CONCLUSIONS}\label{sec:conclusion}
\vspace{-0.1cm}

In this study, we release the multi-modal outdoor surveillance dataset, called \textit{X-MAS}, collected by mobile robots and fixed sensor modules during long-term tests of the system over five years. We assure that the presented dataset with high-quality annotations can provide more mobile robot surveillance research opportunities than existing public outdoor datasets. The study presents several use-cases for exploiting the dataset. By cascading presented algorithms for detection, tracking, and action classification, and we applied this pipeline to our surveillance robot that recognizes each person’s actions and situations in unmanned outdoor surveillance.

In addition, the dataset can be used to study context-awareness related to humans in outdoor environments using multi-modal sensors. Context awareness can be conducted using the probability map\footnote[8]{https://github.com/lge-robot-navi/Abnormal-Situation-Detection} generated by accumulating results of detection, tracking, and action recognition.

For future research, efficient fusion of images with different FOVs will be considered. The presented dataset is managed by KIRO (Korea Institute of Robotics and Technology Convergence) and can be downloaded according to the guideline described in Appendix~\ref{sec:How to Download Dataset}.

\appendix
\section{APPENDIX}

\subsection{Sensor Specifications}\label{sec:Sensor Specifications}
\vspace{-0.5cm}
\begin{table}[h]
\begin{center}
	\setlength{\tabcolsep}{3pt}
\begin{tabular}{c|l|l|c}
\toprule \midrule
 \scriptsize Sensor & \makecell{\scriptsize Part number} & \makecell{\scriptsize FOV} & \makecell{\scriptsize FPS (Hz)} \\ \midrule
  \scriptsize RGB-D &  \scriptsize Realsense D435i (Intel) &  \scriptsize 85.2°(H) $\times$ 58°(V) & \scriptsize 30\\
  \scriptsize RGB  &  \scriptsize Mars640-300GC (Contrastech) &  \scriptsize 69.4°(H) $\times$ 42.5°(V) & \scriptsize 300\\
  \scriptsize Thermal  &   \scriptsize A65 (FLIR) &  \scriptsize 90°(H) $\times$ 69°(V)  & \scriptsize 30\\
  \scriptsize NightVision   &  \scriptsize Night Eagle 2 Pro (Funcam) &  \scriptsize 116.8°(H) $\times$ 101.3°(V) & \scriptsize 38\\
  \scriptsize 3D LiDAR  &  \scriptsize VLP-16 (Velodyne) &   \scriptsize 360°(H) $\times$ 30°(V) & \scriptsize 10\\
 \bottomrule \midrule
\end{tabular}
\end{center}
\vspace{-0.8cm}
\end{table}

\subsection{Characteristics of Each Dataset}\label{sec:Characteristics of Each Dataset}
\vspace{-0.5cm}
\begin{table}[h]
\centering
\setlength{\tabcolsep}{3pt}
\begin{tabular}{c|c|c|c|c}
\toprule \midrule
\scriptsize Data type&  \makecell{\scriptsize Multi-modal}& \makecell{\scriptsize Mask$^{6}$/\\bounding box$^{7}$}& \makecell{\scriptsize Tracking\\scenario}   & \makecell{\scriptsize Tracking ID}  \\
 \midrule
 \scriptsize \textit{X-MAS}-A & \checkmark & \checkmark &  &  \\
 \scriptsize \textit{X-MAS}-B & \checkmark & \checkmark & \checkmark &  \\
 \scriptsize \textit{X-MAS}-C & \checkmark & \checkmark & \checkmark & \checkmark \\
 \bottomrule \midrule
\end{tabular}
\vspace{-0.4cm}
\end{table}
\scriptsize{$^{6}$Annotation for action classification DB, $^{7}$Annotation for detection/tracking DB}
\vspace{-0.4cm}

\normalsize

\
\vspace{-0.3cm}
\subsection{Description of the Reduced Dataset}\label{sec:Reduced Dataset}
\small{The reduced dataset is constructed  using 1-01d, 1-02d, 1-05d, 1-06d, 1-07d, 1-10d, 2-01d, 2-02d, MV\_indoor, MV\_indoor2, and MV\_outdoor\_night as a training/validation dataset.}

\vspace{-0.25cm}
\subsection{How to Download Dataset}\label{sec:How to Download Dataset}

\vspace{-0.1cm}
If you send the purpose, name (including affiliation), and e-mail address to the author (uty@kiro.re.kr), the password for the download website will be delivered. The website is \small{\url{http://gofile.me/6GfMG/eYjbJSjvF}}.

\section*{ACKNOWLEDGMENT}
We would like to thank the numerous prominent researchers of  LG Electronics, KIRO, SNU, KAIST, and ETRI for their hard work in developing, evaluating, and constructing datasets over the past five years. 
\vspace{-0.15cm}


\bibliographystyle{IEEEtran}
\bibliography{./references,./IEEEabrv}

\begin{thebibliography}{10}
\providecommand{\url}[1]{#1}
\csname url@rmstyle\endcsname
\providecommand{\newblock}{\relax}
\providecommand{\bibinfo}[2]{#2}
\providecommand\BIBentrySTDinterwordspacing{\spaceskip=0pt\relax}
\providecommand\BIBentryALTinterwordstretchfactor{4}
\providecommand\BIBentryALTinterwordspacing{\spaceskip=\fontdimen2\font plus
\BIBentryALTinterwordstretchfactor\fontdimen3\font minus
  \fontdimen4\font\relax}
\providecommand\BIBforeignlanguage[2]{{%
\expandafter\ifx\csname l@#1\endcsname\relax
\typeout{** WARNING: IEEEtran.bst: No hyphenation pattern has been}%
\typeout{** loaded for the language `#1'. Using the pattern for}%
\typeout{** the default language instead.}%
\else
\language=\csname l@#1\endcsname
\fi
#2}}

\bibitem{ginting2021chord}
M.~F. Ginting, K.~Otsu, J.~A. Edlund, J.~Gao, and A.-A. Agha-Mohammadi,
  ``{CHORD}: Distributed data-sharing via hybrid {ROS} 1 and 2 for multi-robot
  exploration of large-scale complex environments,'' \emph{IEEE Robot. Automat.
  Lett.}, vol.~6, no.~3, pp. 5064--5071, 2021.

\bibitem{zaheer2021anomaly}
M.~Z. Zaheer, A.~Mahmood, M.~H. Khan, M.~Astrid, and S.-I. Lee, ``An anomaly
  detection system via moving surveillance robots with human collaboration,''
  in \emph{Proc. IEEE/CVF Int. Conf. on Comput. Vis. (ICCV)}, 2021, pp.
  2595--2601.

\bibitem{hoshino2015probabilistic}
S.~Hoshino and T.~Ishiwata, ``Probabilistic surveillance by mobile robot for
  unknown intruders,'' in \emph{Proc. IEEE/RSJ Int. Conf. Intell. Robot. Syst.
  (IROS)}, 2015, pp. 623--629.

\bibitem{xu2010systems}
Y.~Xu and D.~Song, ``Systems and algorithms for autonomous and scalable crowd
  surveillance using robotic {PTZ} cameras assisted by a wide-angle camera,''
  \emph{Autonomous Robots}, vol.~29, no.~1, pp. 53--66, 2010.

\bibitem{chen2019suma++}
X.~Chen, A.~Milioto, E.~Palazzolo, P.~Giguere, J.~Behley, and C.~Stachniss,
  ``Suma++: Efficient {LiDAR}-based semantic slam,'' in \emph{Proc. IEEE/RSJ
  Int. Conf. Intell. Robot. Syst. (IROS)}, 2019, pp. 4530--4537.

\bibitem{kim2022step}
Y.~Kim, B.~Yu, E.~M. Lee, J.-H. Kim, H.-W. Park, and H.~Myung, ``{STEP}: State
  estimator for legged robots using a preintegrated foot velocity factor,''
  \emph{IEEE Robot. Automat. Lett.}, vol.~7, no.~2, pp. 4456--4463, 2022.

\bibitem{lim2021patchwork}
H.~Lim, M.~Oh, and H.~Myung, ``Patchwork: Concentric zone-based region-wise
  ground segmentation with ground likelihood estimation using a 3{D} {LiDAR}
  sensor,'' \emph{IEEE Robot. Automat. Lett.}, vol.~6, no.~4, pp. 6458--6465,
  2021.

\bibitem{bozcan2021gridnet}
I.~Bozcan, J.~Le~Fevre, H.~X. Pham, and E.~Kayacan, ``Grid{N}et: Image-agnostic
  conditional anomaly detection for indoor surveillance,'' \emph{IEEE Robot.
  Automat. Lett.}, vol.~6, no.~2, pp. 1638--1645, 2021.

\bibitem{khaire2022semi}
P.~Khaire and P.~Kumar, ``A semi-supervised deep learning based video anomaly
  detection framework using {RGB-D} for surveillance of real-world critical
  environments,'' \emph{Elsevier Journal of Forensic Science International:
  Digital Investigation}, vol.~40, p. 301346, 2022.

\bibitem{gruosso2021human}
M.~Gruosso, N.~Capece, and U.~Erra, ``Human segmentation in surveillance video
  with deep learning,'' \emph{Journal of Multimedia Tools and Applications},
  vol.~80, no.~1, pp. 1175--1199, 2021.

\bibitem{bozcan2020uav}
I.~Bozcan and E.~Kayacan, ``{UAV-AdNet}: Unsupervised anomaly detection using
  deep neural networks for aerial surveillance,'' in \emph{Proc. IEEE/RSJ Int.
  Conf. Intell. Robot. Syst. (IROS)}, 2020, pp. 1158--1164.

\bibitem{Rudenko2021Learning}
A.~Rudenko, L.~Palmieri, J.~Doellinger, A.~J. Lilienthal, and K.~O. Arras,
  ``Learning occupancy priors of human motion from semantic maps of urban
  environments,'' \emph{IEEE Robot. Automat. Lett.}, vol.~6, no.~2, pp.
  3248--3255, 2021.

\bibitem{liu2019ntu}
J.~Liu, A.~Shahroudy, M.~Perez, G.~Wang, L.-Y. Duan, and A.~C. Kot, ``{NTU
  RGB+D} 120: A large-scale benchmark for 3{D} human activity understanding,''
  \emph{IEEE Trans. on Pattern Anal. Mach. Intell. (PAMI)}, vol.~42, no.~10,
  pp. 2684--2701, 2019.

\bibitem{Ye9696362Tracker}
J.~Ye, C.~Fu, Z.~Cao, S.~An, G.~Zheng, and B.~Li, ``Tracker meets night: A
  transformer enhancer for {UAV} tracking,'' \emph{IEEE Robot. Automat. Lett.},
  vol.~7, no.~2, pp. 3866--3873, 2022.

\bibitem{lei2006real}
B.~Lei and L.-Q. Xu, ``Real-time outdoor video surveillance with robust
  foreground extraction and object tracking via multi-state transition
  management,'' \emph{Pattern Recognit. Lett.}, vol.~27, no.~15, pp.
  1816--1825, 2006.

\bibitem{vertens2020heatnet}
J.~Vertens, J.~Z{\"u}rn, and W.~Burgard, ``Heat{N}et: Bridging the day-night
  domain gap in semantic segmentation with thermal images,'' in \emph{Proc.
  IEEE/RSJ Int. Conf. Intell. Robot. Syst. (IROS)}, 2020, pp. 8461--8468.

\bibitem{choi2016thermal}
Y.~Choi, N.~Kim, S.~Hwang, and I.~S. Kweon, ``Thermal image enhancement using
  convolutional neural network,'' in \emph{Proc. IEEE/RSJ Int. Conf. Intell.
  Robot. Syst. (IROS)}, 2016, pp. 223--230.

\bibitem{9664374}
J.~Yin, A.~Li, T.~Li, W.~Yu, and D.~Zou, ``{M2DGR}: A multi-sensor and
  multi-scenario {SLAM} dataset for ground robots,'' \emph{IEEE Robot. Automat.
  Lett.}, vol.~7, no.~2, pp. 2266--2273, 2022.

\bibitem{sultani2018real}
W.~Sultani, C.~Chen, and M.~Shah, ``Real-world anomaly detection in
  surveillance videos,'' in \emph{Proc. IEEE/CVF Conf. Comput. Vis. Pattern
  Recognit. (CVPR)}, 2018, pp. 6479--6488.

\bibitem{baltieri20113dpes}
D.~Baltieri, R.~Vezzani, and R.~Cucchiara, ``3{DP}e{S}: 3{D} people dataset for
  surveillance and forensics,'' in \emph{Proc. Joint ACM Workshop on Human
  Gesture and Behavior Understanding}, 2011, pp. 59--64.

\bibitem{lim2021deep}
J.~Lim, M.~I. Al~Jobayer, V.~M. Baskaran, J.~M. Lim, J.~See, and K.~Wong,
  ``Deep multi-level feature pyramids: Application for non-canonical firearm
  detection in video surveillance,'' \emph{Journal of Engineering Applications
  of Artificial Intelligence}, vol.~97, p. 104094, 2021.

\bibitem{naphade20182018}
M.~Naphade, M.-C. Chang, A.~Sharma, D.~C. Anastasiu, V.~Jagarlamudi,
  P.~Chakraborty, T.~Huang, S.~Wang, M.-Y. Liu, R.~Chellappa, \emph{et~al.},
  ``The 2018 {NVIDIA AI} city challenge,'' in \emph{Proc. IEEE/CVF Conf.
  Comput. Vis. Pattern Recognit. Workshops (CVPR)}, 2018, pp. 53--60.

\bibitem{choi2009they}
W.~Choi, K.~Shahid, and S.~Savarese, ``What are they doing?: Collective
  activity classification using spatio-temporal relationship among people,'' in
  \emph{Proc. IEEE/CVF Int. Conf. Comput. Vis. Workshops (ICCV)}, 2009, pp.
  1282--1289.

\bibitem{coppola2016social}
C.~Coppola, D.~R. Faria, U.~Nunes, and N.~Bellotto, ``Social activity
  recognition based on probabilistic merging of skeleton features with
  proximity priors from {RGB-D} data,'' in \emph{Proc. IEEE/RSJ Int. Conf.
  Intell. Robot. Syst. (IROS)}, 2016, pp. 5055--5061.

\bibitem{russakovsky2015imagenet}
O.~Russakovsky, J.~Deng, H.~Su, J.~Krause, S.~Satheesh, S.~Ma, Z.~Huang,
  A.~Karpathy, A.~Khosla, M.~Bernstein, \emph{et~al.}, ``Image{N}et large scale
  visual recognition challenge,'' \emph{Int. J. Comput. Vis.}, vol. 115, no.~3,
  pp. 211--252, 2015.

\bibitem{everingham2015pascal}
M.~Everingham, S.~A. Eslami, L.~Van~Gool, C.~K. Williams, J.~Winn, and
  A.~Zisserman, ``The {PASCAL} visual object classes challenge: A
  retrospective,'' \emph{Int. J. Comput. Vis.}, vol. 111, no.~1, pp. 98--136,
  2015.

\bibitem{lin2014microsoft}
T.-Y. Lin, M.~Maire, S.~Belongie, J.~Hays, P.~Perona, D.~Ramanan,
  P.~Doll{\'a}r, and C.~L. Zitnick, ``Microsoft {COCO}: Common objects in
  context,'' in \emph{Proc. European Conf. Comput. Vis. (ECCV)}, 2014, pp.
  740--755.

\bibitem{baktashmotlagh2013unsupervised}
M.~Baktashmotlagh, M.~T. Harandi, B.~C. Lovell, and M.~Salzmann, ``Unsupervised
  domain adaptation by domain invariant projection,'' in \emph{Proc. IEEE/CVF
  Int. Conf. Comput. Vis. (ICCV)}, 2013, pp. 769--776.

\bibitem{aytar2011tabula}
Y.~Aytar and A.~Zisserman, ``Tabula {R}asa: Model transfer for object category
  detection,'' in \emph{Proc. IEEE/CVF Int. Conf. Comput. Vis. (ICCV)}, 2011,
  pp. 2252--2259.

\bibitem{zaheer2020self}
M.~Z. Zaheer, A.~Mahmood, H.~Shin, and S.-I. Lee, ``A self-reasoning framework
  for anomaly detection using video-level labels,'' \emph{IEEE Signal
  Processing Letters}, vol.~27, pp. 1705--1709, 2020.

\bibitem{bhardwaj2021computationally}
R.~Bhardwaj, A.~Dhull, and M.~Sharma, ``A computationally efficient real-time
  vehicle and speed detection system for video traffic surveillance,'' in
  \emph{Proc. International Conference on Artificial Intelligence and
  Applications}, 2021, pp. 583--594.

\bibitem{9760091}
A.~J. Lee, Y.~Cho, Y.-S. Shin, A.~Kim, and H.~Myung, ``{ViViD++}: Vision for
  visibility dataset,'' \emph{IEEE Robot. Automat. Lett.}, vol.~7, no.~3, pp.
  6282--6289, 2022.

\bibitem{feng2020deep}
D.~Feng, C.~Haase-Sch{\"u}tz, L.~Rosenbaum, H.~Hertlein, C.~Glaeser, F.~Timm,
  W.~Wiesbeck, and K.~Dietmayer, ``Deep multi-modal object detection and
  semantic segmentation for autonomous driving: Datasets, methods, and
  challenges,'' \emph{IEEE Trans. Intell. Transport. Syst. (ITS)}, vol.~22,
  no.~3, pp. 1341--1360, 2020.

\bibitem{electronics11142214}
T.~Uhm, J.~Park, J.~Lee, G.~Bae, G.~Ki, and Y.~Choi, ``Design of multimodal
  sensor module for outdoor robot surveillance system,'' \emph{Electronics},
  vol.~11, no.~14, p. 2214, 2022.

\bibitem{https://doi.org/10.4218/etrij.2021-0395}
H.~Shin, K.-I. Na, J.~Chang, and T.~Uhm, ``Multimodal layer surveillance map
  based on anomaly detection using multi-agents for smart city security,''
  \emph{ETRI Journal}, vol.~44, no.~2, pp. 183--193, 2022.

\bibitem{wang2014cdnet}
Y.~Wang, P.-M. Jodoin, F.~Porikli, J.~Konrad, Y.~Benezeth, and P.~Ishwar,
  ``{CDnet} 2014: An expanded change detection benchmark dataset,'' in
  \emph{Proc. IEEE/CVF Conf. Comput. Vis. Pattern Recognit. (CVPR) Workshops},
  2014, pp. 387--394.

\bibitem{chen2018encoder}
L.-C. Chen, Y.~Zhu, G.~Papandreou, F.~Schroff, and H.~Adam, ``Encoder-decoder
  with atrous separable convolution for semantic image segmentation,'' in
  \emph{Proc. European Conf. Comput. Vis. (ECCV)}, 2018, pp. 801--818.

\bibitem{wu2013online}
Y.~Wu, J.~Lim, and M.-H. Yang, ``Online object tracking: A benchmark,'' in
  \emph{Proc. IEEE/CVF Conf. Comput. Vis. and Pattern Recognit. (CVPR)}, 2013,
  pp. 2411--2418.

\bibitem{fan2019lasot}
H.~Fan, L.~Lin, F.~Yang, P.~Chu, G.~Deng, S.~Yu, H.~Bai, Y.~Xu, C.~Liao, and
  H.~Ling, ``La{SOT}: A high-quality benchmark for large-scale single object
  tracking,'' in \emph{Proc. IEEE/CVF Conf. Comput. Vis. and Pattern Recognit.
  (CVPR)}, 2019, pp. 5374--5383.

\bibitem{muller2018trackingnet}
M.~Muller, A.~Bibi, S.~Giancola, S.~Alsubaihi, and B.~Ghanem, ``Tracking{N}et:
  A large-scale dataset and benchmark for object tracking in the wild,'' in
  \emph{Proc. European Conf. Comput. Vis. (ECCV)}, 2018, pp. 300--317.

\bibitem{huang2019got}
L.~Huang, X.~Zhao, and K.~Huang, ``{GOT}-10k: A large high-diversity benchmark
  for generic object tracking in the wild,'' \emph{IEEE Trans. Pattern Anal.
  Mach. Intell. (PAMI)}, vol.~43, no.~5, pp. 1562--1577, 2019.

\bibitem{7533003}
A.~Bewley, Z.~Ge, L.~Ott, F.~Ramos, and B.~Upcroft, ``Simple online and
  realtime tracking,'' in \emph{Proc. IEEE International Conference on Image
  Processing (ICIP)}, 2016, pp. 3464--3468.

\bibitem{kalman1960new}
\BIBentryALTinterwordspacing
R.~E. Kalman, ``A new approach to linear filtering and prediction problems,''
  \emph{ASME. Trans. J. Basic Eng.}, vol.~82, no.~1, pp. 35--45, Mar. 1960.
  [Online]. Available: \url{https://doi.org/10.1115/1.3662552}
\BIBentrySTDinterwordspacing

\bibitem{kuhn1955hungarian}
H.~W. Kuhn, ``The hungarian method for the assignment problem,'' \emph{Naval
  Research Logistics Quarterly}, vol.~2, no. 1-2, pp. 83--97, 1955.

\bibitem{milan2016mot16}
A.~Milan, L.~Leal-Taix{\'e}, I.~Reid, S.~Roth, and K.~Schindler, ``{MOT}16: A
  benchmark for multi-object tracking,'' \emph{arXiv preprint
  arXiv:1603.00831}, 2016.

\bibitem{soomro2012ucf101}
K.~Soomro, A.~R. Zamir, and M.~Shah, ``{UCF101}: A dataset of 101 human actions
  classes from videos in the wild,'' \emph{arXiv preprint arXiv:1212.0402},
  2012.

\bibitem{he2016deep}
K.~He, X.~Zhang, S.~Ren, and J.~Sun, ``Deep residual learning for image
  recognition,'' in \emph{Proc. IEEE/CVF Conf. Comput. Vis. Pattern Recognit.
  (CVPR)}, 2016, pp. 770--778.

\end{thebibliography}

\addtolength{\textheight}{-12cm}   


\end{document}